\documentclass[conference, anonymous, review]{IEEEtran}
\IEEEoverridecommandlockouts

\usepackage{cite}
\usepackage{amsmath,amssymb,amsfonts}
\usepackage{algorithmic}
\usepackage{graphicx}
\usepackage{textcomp}
\usepackage{xcolor}
\usepackage{booktabs}
\usepackage{multirow}
\usepackage{float}
\usepackage{subcaption}
\usepackage{adjustbox}
\usepackage{caption} 
\usepackage{url}
\usepackage{enumitem}
\usepackage{makecell}

\def\BibTeX{{\rm B\kern-.05em{\sc i\kern-.025em b}\kern-.08em
    T\kern-.1667em\lower.7ex\hbox{E}\kern-.125emX}}

\begin{document}

\title{Automated Knot Detection and Pairing for Wood Analysis in the Timber Industry}

\author{
\IEEEauthorblockN{
Guohao Lin\IEEEauthorrefmark{1},
Shidong Pan\IEEEauthorrefmark{2},
Rasul Khanbayov\IEEEauthorrefmark{1},\\
Changxi Yang\IEEEauthorrefmark{1},
Ani Khaloian-Sarnaghi\IEEEauthorrefmark{1},
Andriy Kovryga\IEEEauthorrefmark{3}
}
\IEEEauthorblockA{\IEEEauthorrefmark{1}Technical University of Munich, Germany \\
\{guohao.lin, rasul.khanbayov, changxi.yang\}@tum.de}
\IEEEauthorblockA{\IEEEauthorrefmark{2}Australian National University \& CSIRO's Data61, Australia \\
shidong.pan@anu.edu.au}
\IEEEauthorblockA{\IEEEauthorrefmark{3}Hochschule für Musik und Theater München, Germany \\
\{khaloian, kovryga\}@hfm.tum.de}
}

\maketitle

\begin{abstract}
Knots in wood are critical to both aesthetics and structural integrity, making their detection and pairing essential in timber processing. 
However, traditional manual annotation was labor-intensive and inefficient, necessitating automation.
This paper proposes a lightweight and fully automated pipeline for knot detection and pairing based on machine learning techniques. 
In the detection stage, high-resolution surface images of wooden boards were collected using industrial-grade cameras, and a large-scale dataset was manually annotated and preprocessed. 
After the transfer learning, the YOLOv8l achieves an mAP@0.5 of 0.887.
In the pairing stage, detected knots were analyzed and paired based on multidimensional feature extraction. A triplet neural network was used to map the features into a latent space, enabling clustering algorithms to identify and pair corresponding knots. The triplet network with learnable weights achieved a pairing accuracy of 0.85. Further analysis revealed that he distances from the knot's start and end points to the bottom of the wooden board, and the longitudinal coordinates play crucial roles in achieving high pairing accuracy.
Our experiments validate the effectiveness of the proposed solution, demonstrating the potential of AI in advancing wood science and industry.
\end{abstract}

\section{Introduction}
\label{sec:introduction}
Knots, as natural defects in wood, play a crucial role in determining timber's structural integrity, mechanical performance, and overall quality. Their presence reduces wood's usability, strength, and durability, while also introducing challenges such as aesthetic inconsistency and susceptibility to damage \cite{foley2003modeling,xu2002estimating,qu2020effect,boatright1979effect,as2006effect,guindos2014analytical}. Knots differ significantly from the surrounding wood tissue in terms of density and water content, which can result in structural weaknesses, such as holes forming during the drying process \cite{jin2016discussion}. It has been reported that knots can account for up to 70\% of downgrading among all wood defects \cite{thelandersson2003timber}. Accurately identifying and analyzing knots is therefore essential for optimizing cutting plans, improving aesthetics, and ensuring timber strength.
In addition, the knot pairing task is significant in the wood industry as it helps assess the structural integrity and aesthetic quality of lumber. By identifying and analyzing pairs of knots, manufacturers can predict weak points in the wood, optimize cutting strategies, and improve material selection for different applications.

Traditionally, knot detection and pairing have been performed through manual inspection or advanced imaging technologies, such as X-ray scanning \cite{longuetaud2012automatic,krahenbuhl2012knot,pietikainen1996detection,omori2023log,stangle2015potentially}. While effective, these methods are often time-consuming, labor-intensive, and reliant on costly equipment, making them impractical for large-scale industrial applications. To address these limitations, this study proposes a lightweight and fully automated pipeline for knot detection and pairing that leverages machine learning techniques based on neural networks.

The proposed approach consists of two primary stages: knot detection and knot pairing.
In the detection stage, we collected high-resolution surface images of wooden boards using industrial-grade cameras. A large-scale dataset was manually annotated and fine-grained preprocessing was applied to ensure data quality. The dataset was then used to train three YOLOv8 variants via transfer learning~\cite{torrey2010transfer, weiss2016survey, zhuang2020comprehensive}. Among these models, YOLOv8l achieved an mAP@0.5 of 0.887, while the lightweight YOLOv8n achieved 0.846 mAP@0.5, with only 1/20 of the model size and 1/8 of the training time.

Following detection, the pairing stage analyzes the detected knots and pairs them based on extracted multidimensional knot features. These features, which are easily derived from continuous images, are preprocessed through normalization and cleaning to ensure consistency~\cite{albahra2023artificial}. A triplet neural network\cite{ren2024exploring} is then employed to map the features into a latent space, where clustering algorithms\cite{blashfield1978literature, wu2012cluster} identify and pair corresponding knots. This step also enables an exploration of how different knot characteristics influence the pairing process, providing insights for wood science. The triplet network with learnable weights achieved a pairing accuracy of 0.85. Further analysis reveals that the features k1 (the distances from the knot's start point to the bottom of the wooden board), k2 (the distances from the knot's end point to the bottom of the wooden board), and longitudinal coordinates play crucial roles in achieving high pairing accuracy.

By combining computer vision-based detection with feature-driven pairing methodologies, this work introduces a scalable and efficient solution for automating knot analysis in timber processing. The results demonstrate the feasibility of replacing manual inspections with a robust pipeline, which can significantly enhance the speed and accuracy of knot detection and pairing. This study also demonstrates the feasibility of applying AI techniques to wood science and industry, exhibiting its interdisciplinary impact.

\section{Background and Related Work}
\label{sec:background}
\subsection{Knots of Wood }
\begin{figure}[t]
  \centering
    \includegraphics[width=0.47\textwidth]{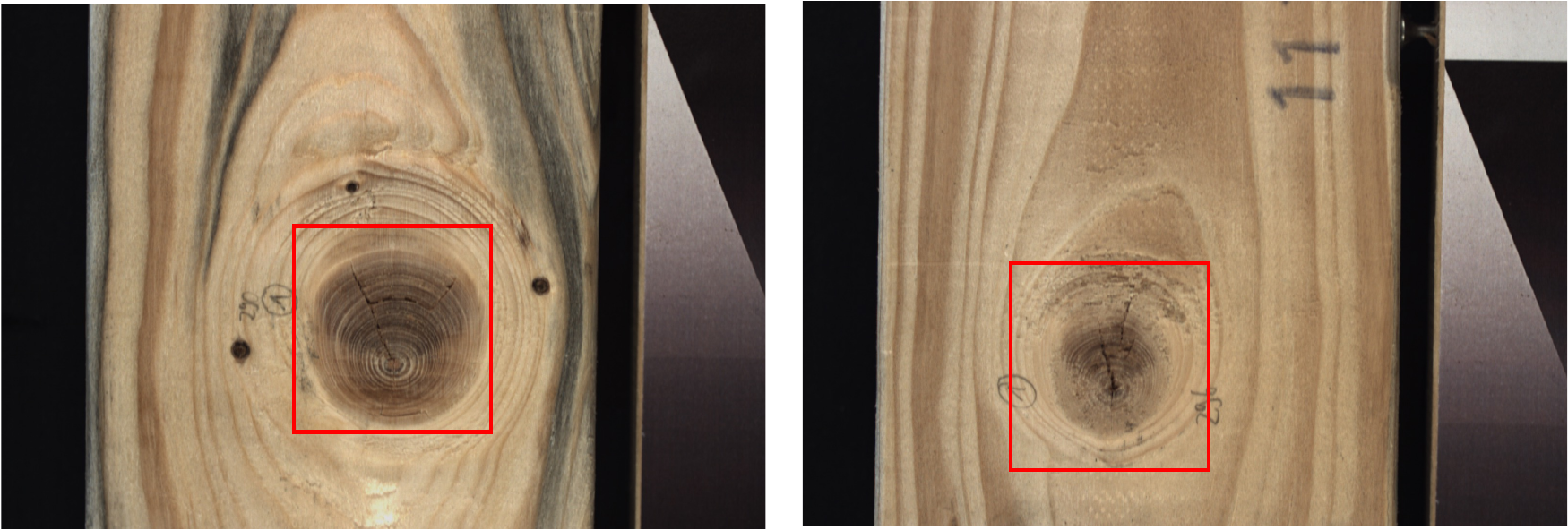}
    \caption{Two examples of knots on the different surfaces of the same wooden board. They are paired to each other.}
    \label{fig:knots_example}
\end{figure}

In timber and wood science, a knot is the visible remnant of a tree's branching process, appearing as a circular or oval imperfection embedded in the trunk or limbs \cite{fpl1987wood}. Figure~\ref{fig:knots_example} illustrates two examples of a knot (marked in red). Knots occur naturally when subsequent trunk growth layers enclose a branch. They vary in size, color, and density, often appearing darker than the surrounding wood \cite{qu2020effect}. Depending on the different types and sizes, knots can reduce the cross sectional area and influence the fiber orientation around the knot differently, resulting in strength reduction to different extents. Knots can also cause stress concentration and may initiate failure at their adjacent locations much earlier than in sound wood. The strength prediction of timber engineering products generally involves knot assessment, which can be accomplished through surface scanning systems or X-ray systems \cite{khaloian2019advanced}. 

\subsection{Object Detection in Wood Science}
Object detection and segmentation are crucial in wood science for identifying defects such as knots, cracks, and grain irregularities~\cite{funck2003image,yang2020wood}. 
Unlike general object detection tasks, detecting knots poses unique challenges due to the low contrast, uniform color, and minimal texture variation of wood images, which are typically two-dimensional and captured under controlled conditions. These characteristics result in low information density, making it harder for models to distinguish knots from the background.
YOLO (You Only Look Once) is the fast and precise state-of-the-art object detection model, but its performance suffers on wood images where knots are small, irregular, and lack distinct features~\cite{jiang2022review,diwan2023object}. 
Other models, such as DeepLabV3~\cite{chen2017rethinking} also struggle with the subtle contrasts and patterns of wood surfaces. 
Moreover, both models face limitations in industrial applications requiring real-time processing.
YOLO's efficiency drops with fine-grained detection, and DeepLabV3's high computational demands make it impractical for fast workflows.

\subsection{Knots Pairing}~\label{sec_back_pair}
\begin{figure}[!t]
  \centering
  \includegraphics[width=0.4\textwidth]{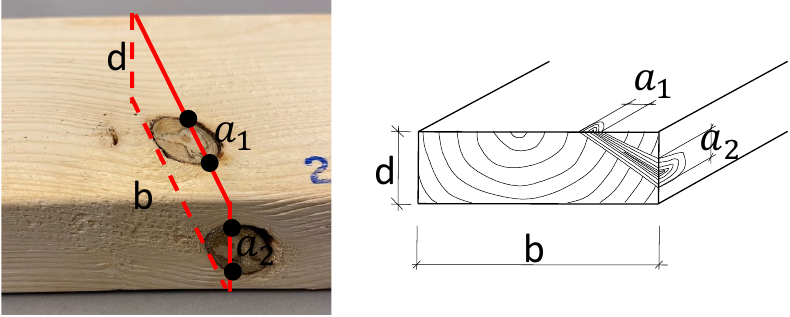}
  \caption{An illustration of knot pairing on a wooden board. The image to the left is a real-world photograph showing the actual knot pair on a wooden board. The diagram to the right is a cross-section that shows how the two knots are interconnected, forming a pair that appears on adjacent surfaces of the board.}
  \label{fig:knots_pairing}
\end{figure}

Knot pairing refers to the process of identifying and matching knots across different surfaces of a wooden board to establish their spatial and geometric relationships. As shown in Figure~\ref{fig:knots_pairing}, a knot inside the wooden board can appear on multiple surfaces due to its three-dimensional structure. For instance, Knot A and Knot B are visible on adjacent surfaces. These knots represent different views of the same physical feature within the board.

In general, a single knot may be visible on up to four surfaces of a wooden board. The task of knot pairing involves grouping such related knots together. Only knots that represent the same physical entity can be considered part of a "knot pair." Correct identification of knot pairs helps improve the accuracy of strength prediction of timber engineering products. Up to now, knot pairing largely relies on the manual visual grading process. Experienced workers determine the knot pairs based on the spatial alignment, geometric continuity, and relative locations to the pith. For a 2-meter-long, 0.5-meter-wide wooden board, dozens of knots may need to be identified and paired. This manual annotation process is highly inefficient, labor-intensive, and prone to human error, calling for the need for automated solutions.

Effective detection and pairing in wood science requires analyzing the spatial relationships of knots across board surfaces, which general models are not designed to handle. 
These challenges emphasize the need for lightweight, specialized models optimized for the unique properties of wood images, leveraging tailored preprocessing and transfer learning to balance accuracy and speed for industrial deployment.

\section{Methodology}
\begin{figure*}[ht]
    \centering
    \includegraphics[width=0.95\textwidth]{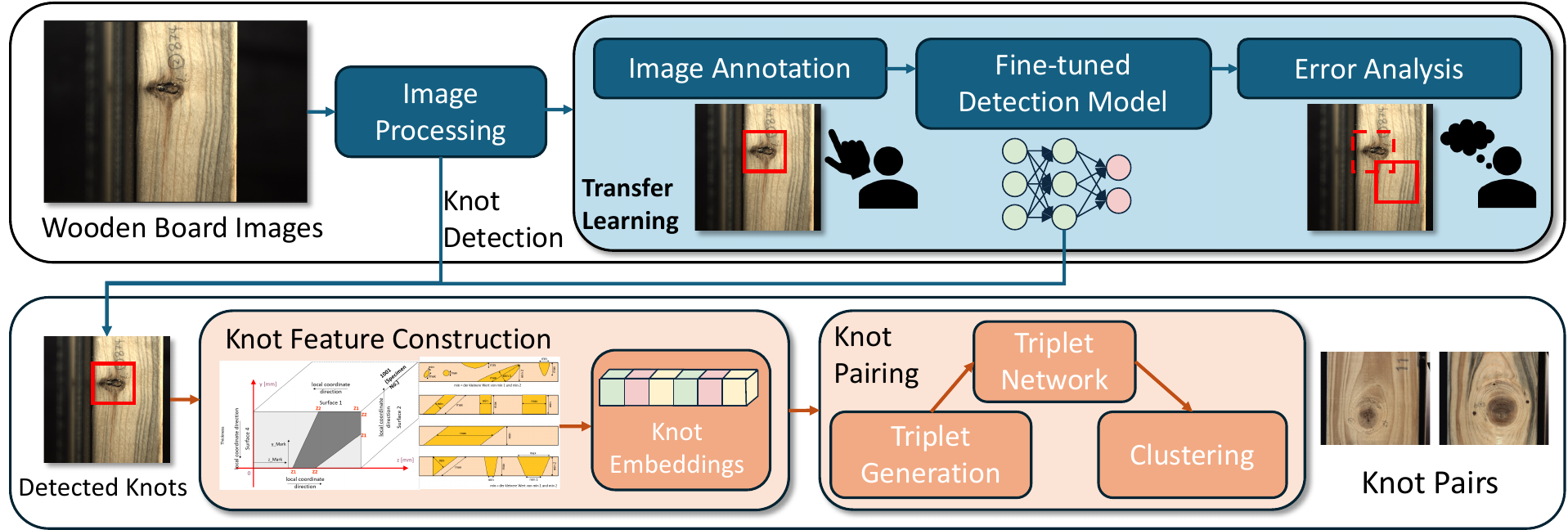}
    \caption{The proposed pipeline for knots detection and pairing. The workflow consists of two main stages: Knots Detection (the top section) and Knots Pairing (the bottom section).}
    \label{fig:pipeline}
\end{figure*}

The proposed methodology aims to streamline the process of detecting and pairing knots in wooden boards, addressing the challenges of manual labor and inefficiencies in traditional methods. Figure~\ref{fig:pipeline} shows the pipeline consists of two primary stages: Knot Detection and Knot Pairing.

%


\subsection{Knot Detection}
In the knot detection stage, we first batch-processed the original wooden board images pictured by factory cameras, and then trained state-of-the-art object detection models through transfer learning to identify knots on wooden boards. 

\subsubsection{Image Processing}
To ensure a substantial amount of high-resolution data to support our work, we utilized industrial-grade cameras to capture images of wooden boards. Each board was documented across four surfaces (top, bottom, front, and back), with images captured sequentially along a conveyor belt to ensure comprehensive coverage. Although slight overlaps were unavoidable, this structured and detailed dataset provides a solid foundation for the knot detection task.
In total, we obtained 44,735 high-resolution images of 153 spruce boards, and the size is larger than 300GB.
Due to industry requirements, the dataset will be made available upon request after publication.

\begin{figure}[t]
\centering
  \begin{subfigure}[b]{0.2\textwidth} 
  \centering
    \includegraphics[width=\textwidth]{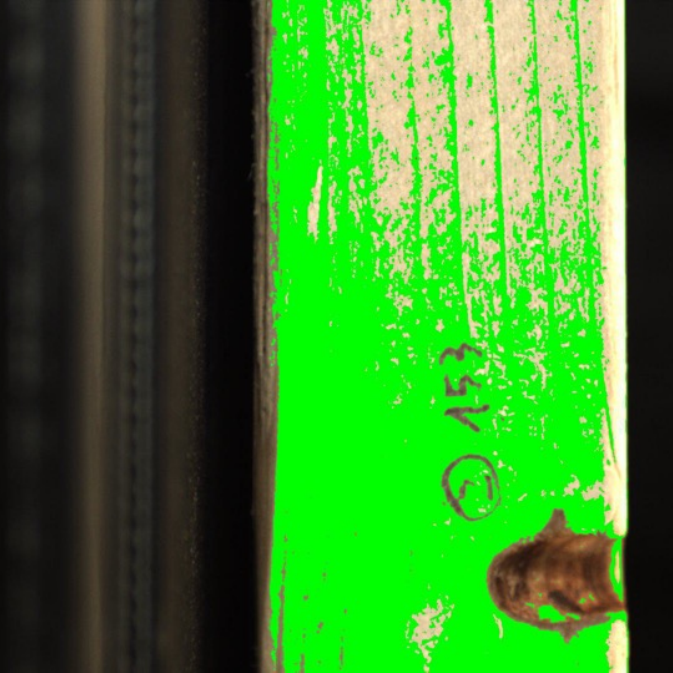}
    \caption{Matched pixels.}
    \label{fig:color_based_filter}
  \end{subfigure}
  \hspace{5pt}
  \begin{subfigure}[b]{0.2\textwidth} 
  \centering
    \includegraphics[width=\textwidth]{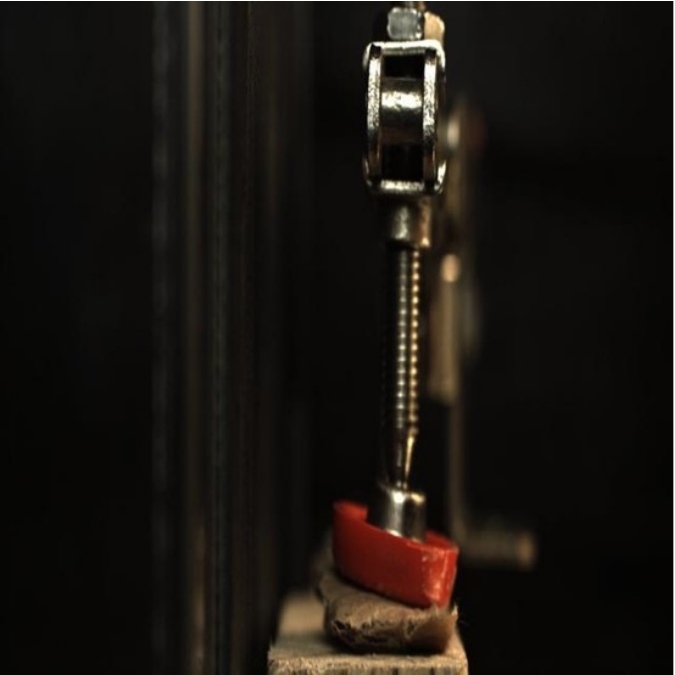}
    \caption{An outlier image.}
    \label{fig:error_outlier}
  \end{subfigure}
\caption{Example of images from the dataset.}
\label{}
\end{figure}

When wooden boards were placed and removed on the conveyor belt for picturing, imaging might be problematic.
Thus we removed those obvious outliers to increase the quality of dataset\cite{dabov2007image}.
Based on our observation, we noticed that most of the wrong pictures do not contain any wood, but mostly black background.
We developed a color-based filtering technique\cite{kulkarni2012color}.
Specifically, we calculated that the average wood color of wooden materials is [$R=190$, $G=161$, $B=125$] based on a random sample from our dataset.
If an image contains less than 5\% pixels that are close to the wood color ($\pm 5$ for any of the RGB values), then the image is regarded as an outlier and removed.
In total, approximately 20\% of images were removed. 
This approach significantly enhances dataset quality by focusing on relevant wood features. 
Figure~\ref{fig:color_based_filter} illustrates an example in the dataset, where the matching pixels are highlighted in green, and Figure~\ref{fig:error_outlier} shows a common outlier that had been filtered out.

\subsubsection{Transfer Learning}

YOLO series models are widely recognized as state-of-the-art in object detection tasks. They have consistently demonstrated excellent performance in real-life scenarios, such as autonomous driving or surveillance~\cite{sarda2021object,li2022cross,narejo2021weapon,nguyen2021yolo}. 
However, unlike these scenarios, images primarily composed of natural patterns tend to have lower information frequency, making object detection more challenging. To address this, we applied transfer learning for knot detection, establishing a foundation for the subsequent knot pairing task.

Therefore, we manually annotated a knots detection dataset that specifically marked wood knots on wooden boards. 
As the unique characteristics of wood knots, including their variability in size, shape, and texture, necessitate a specialized process to ensure high-quality annotation.

For this task, we utilized the makesense.ai\footnote{\url{https://www.makesense.ai/}} platform, a user-friendly, web-based tool commonly recommended for creating YOLO-compatible annotations. This platform allows researchers to upload images and annotate them with precise bounding boxes that define the location of knots in each image. The resulting annotations are saved in YOLO format, enabling seamless integration into the transfer learning pipeline.
This manual annotation process ensures that the dataset captures essential knot details, providing a strong foundation for training detection models optimized for wood surfaces. Subsequently, we fine-tuned pre-trained YOLO models on the annotated dataset on the knot detection task.

\subsection{Knots Pairing}
\subsubsection{Knots Feature Construction}
\begin{figure}[t]
  \centering
  \begin{minipage}[b]{0.45\textwidth}
    \includegraphics[width=\textwidth]{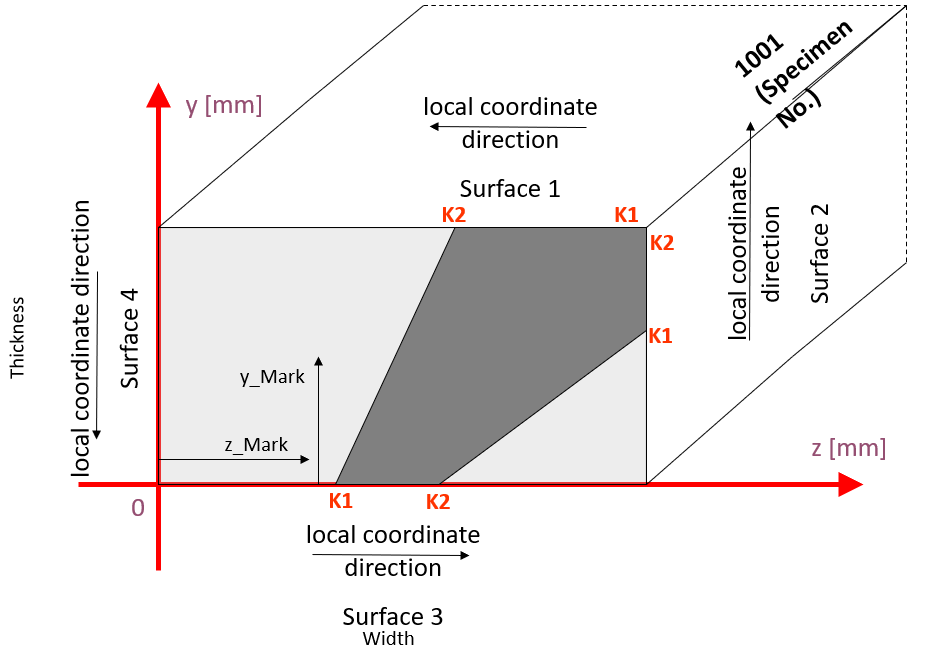}
    \caption{Explaination of `K1' and `K2' values.}
    \label{fig:z1-z2}
  \end{minipage}
\end{figure}

\begin{figure}[t]
  \centering
  \begin{minipage}[b]{0.4\textwidth}
    \includegraphics[width=\textwidth]{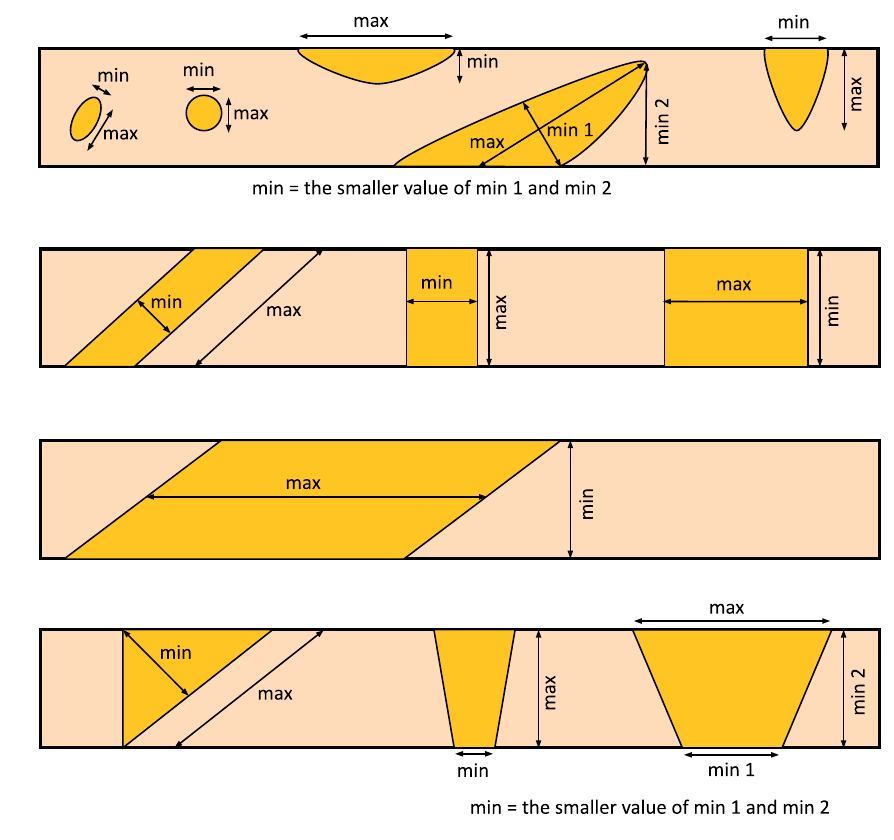}
    \caption{Explanation of `d\_min' and `d\_max' values.}
    \label{fig:min-max}
  \end{minipage}
\end{figure}

\begin{figure}[htbp]
  \centering
  \begin{minipage}[b]{0.5\textwidth}
    \includegraphics[width=\textwidth]{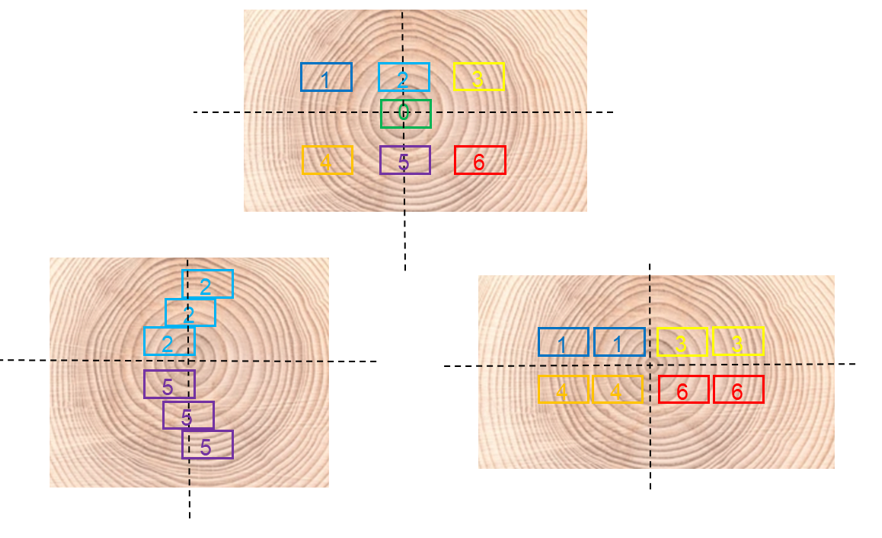}
    \caption{Explanation of Pith Clustering types.}
    \label{fig:pith-clustering}
  \end{minipage}
\end{figure}

As we introduced in Section~\ref{sec_back_pair}, the knot pairing process is a challenging task that requires huge expertise.
After obtaining the location of knots on the wooden board, we can obtain their wooden features through a series of calculations. 
The following features were extracted for each knot:

\begin{itemize} [leftmargin = *]
    \item \textbf{Longitudinal coordinate:} The distance from the beginning of the board to the center of the knot, along the board's longitudinal direction. Assuming the current image is the $n$-th image of the current wooden board, the conveyor belt runs in velocity $v$, and the elapsed time is $t$, then the value can be obtained by $n \times v \times t$, plus the location from knot detection.
    \item \textbf{Surface:} Indicating the knot appearance on the sides of the wooden board, represented by numbers ranging from 1 to 4. Here, 1 indicates the top surface of the board, 2 is the front, 3 is the bottom, and 4 is the back.
    \item \textbf{K1 and K2:} The distances from the bottom of the board to the beginning and end of the knot on that surface, as shown in Figure~\ref{fig:z1-z2}.
    \item \textbf{d\_min and d\_max:} The minimal and maximal dimensions of the knot on that surface, as shown in Figure~\ref{fig:min-max}.
    \item \textbf{Distance of knot center to bottom:} The distance from the knot's center (pith) to the bottom of the corresponding surface. This value is recorded only for surfaces where the pith appears and serves as an approximate radius indicator.
    \item \textbf{Knot type:} A knot can be classified into one of the five categories: Encased knot, Loose knot, Intergrown knot, Overgrown knot, and Dead knot. They are encoded as 1, 2, 3, 4 and 5, respectively.  
    \item \textbf{Pith location:} As visualized in Figure~\ref{fig:pith-clustering}, we use 0 to 6 to reflect the pith's relative location on the cross-section:
    \begin{itemize}
        \item \textbf{L-0:} The board contains the pith.
        \item \textbf{L-1:} The board is upper left to the pith.
        \item \textbf{L-2:} The board is directly above the pith.
        \item \textbf{L-3:} The board is upper right to the pith.
        \item \textbf{L-4:} The board is lower left to the pith.
        \item \textbf{L-5:} The board is directly beneath the pith.
        \item \textbf{L-6:} The board is lower right to the pith.
    \end{itemize}
\end{itemize}

We constructed knot features as embeddings to facilitate the subsequent learning-based approach.
Since many key features in our dataset are relative to the board’s dimensions, they cannot be directly utilized for analysis. Variations in board length, width, and thickness result in inconsistent reference frames, making it difficult for the model to effectively learn from the data.
To address this issue, we performed normalization to standardize feature values. 
The following normalization and processing steps were applied to ensure that each feature contributes equally to the model, regardless of the board’s specific dimensions:

\begin{enumerate}
    \item \textbf{Normalization of Longitudinal Coordinate:}  
    The longitudinal coordinate is normalized based on the length of the board to provide a consistent representation.
    
    \item \textbf{Normalization of Surface Features:}  
    Features such as K1, K2, d\_max, d\_min, and the distance from the knot center to the bottom are normalized differently depending on the surface:
    \begin{enumerate}
        \item For surface numbers 1 or 3, values are normalized by dividing by the width ($b$ in Figure~\ref{fig:knots_pairing}) of the board.
        \item For surface numbers 2 or 4, values are normalized by dividing by the thickness ($d$ in Figure~\ref{fig:knots_pairing}) of the board.
    \end{enumerate}
\end{enumerate}

\subsubsection{Knots Pairing Model}
The triplet network is a type of architecture commonly used for metric learning, where the goal is to learn an embedding space in which similar items are closer together while dissimilar ones are farther apart~\cite{hoffer2015deep}. 
It is widely used in tasks such as face recognition, image retrieval, and object matching~\cite{schroff2015facenet,min2020two,dong2018triplet}.
Therefore, we developed a systematic approach to pair knots based on a triplet network. 

To train the networks effectively, we utilized a triplet-based data preparation approach \cite{xuan2020improved,hoffer2015deep}. For each specimen, the dataset was grouped by unique knot numbers. Positive pairs were created by forming all possible combinations of knots belonging to the same knot class. For each positive pair, a negative sample was selected from a different knot class within the same specimen. This approach ensures that the generated triplets (anchor, positive, negative) represent clear relationships of similarity and dissimilarity between knots, providing the foundation for effective model training.

\begin{itemize} [leftmargin= *]
    \item \textbf{Standard Triplet Network.}  
    The Standard Triplet Network transforms raw knot features into high-dimensional embeddings. Using a triplet loss function \cite{Schroff_2015}, the network minimizes the distance between embeddings of similar knots while maximizing the distance between embeddings of dissimilar knots \cite{kaya2019deep}. This process ensures robust feature representations suitable for downstream clustering tasks.

    \item \textbf{SimCLR-based Method.}  
    The SimCLR-based method applies contrastive learning with extensive data augmentations  \cite{chen2020simple}, such as noise addition, feature scaling, and feature dropping, to improve the resilience of the model\cite{grill2020bootstraplatentnewapproach}. This approach enhances the robustness of feature embeddings, making them less sensitive to variations in the data.

    \item \textbf{Triplet Network with Learnable Weights.}  
    This variation of the Triplet Network introduces learnable weights for each feature, allowing the network to dynamically adjust their importance during training. By optimizing these weights, the network emphasizes the most relevant characteristics of knots, leading to improved accuracy in knot pairing.

    \item \textbf{Triplet Network with Custom Weights.}  
    In this approach, fixed weights based on domain knowledge are assigned to each feature. Features such as "Knot type" and "Pith type" are given higher weights to guide the model toward aspects that significantly influence the accuracy of knot pairing. This approach aims to enhance performance by leveraging prior knowledge and industry-specific intuition\cite{dash2021incorporating,muralidhar2018incorporating}.
\end{itemize}

We then cluster similar knots using feature embeddings generated by trained networks and our clustering approach is based on the principle of Distance Threshold Clustering (DTC)\cite{kanungo2002efficient,10.1145/331499.331504}.
After computing embeddings, pairwise Euclidean distances are calculated to form a distance matrix. A predefined distance threshold is then applied to group knots, ensuring high intra-cluster similarity and low inter-cluster similarity.
By adjusting the distance threshold, the clustering granularity is controlled, balancing precision and recall. This straightforward approach enables effective evaluation of the extracted embeddings and supports real-time integration into automated timber processing workflows.

\section{Experimental Settings}
\subsection{Knot Detection}
\textbf{Knot Detection Dataset.}
The knot detection annotation is a labor-intensive work.
To streamline analysis and support this objective, a dataset of 3,000 images was randomly selected, ensuring the representation of all board surfaces.
After filtering out noisy images, we retained 2,403 images containing visible boards. Among these, 573 images were identified as positive samples with knots manually labeled, while the remaining 1,830 images contained no visible knots. To balance the dataset, 573 images were randomly selected from the negative samples. The final dataset comprised 1,146 images: 573 positive samples and 573 negative samples. The dataset was split into training, validation, and test sets with a ratio of 7:2:1.

\textbf{Model Selection and Training.}  
YOLOv8 is the most advanced model in the YOLO series (at the time we conducted the majority of this research). Three YOLOv8 variants were selected to balance computational efficiency and accuracy:
\begin{itemize}
    \item \textbf{YOLOv8n:} A lightweight model optimized for speed and efficiency, suitable for real-time applications but with lower performance.
    \item \textbf{YOLOv8m:} A medium-sized model offering a balance between computational requirements and performance.
    \item \textbf{YOLOv8l:} A larger model designed for high accuracy, albeit requiring greater computational resources.
\end{itemize}
All models were trained on an NVIDIA GeForce RTX 3050 GPU, with the Adam optimizer. 
The batch sizes were set to 16 for YOLOv8n and YOLOv8m, and 8 for YOLOv8l, reflecting the memory constraints of the RTX 3050. 
The learning rate was initialized at $1 \times 10^{-4}$ and decayed exponentially. Standard data augmentation techniques, including random cropping and horizontal flipping, were applied to enhance generalization.

\textbf{Evaluation Metrics.}  
    Model performance was evaluated using precision, recall, and mean Average Precision (mAP@0.5). Precision measures the accuracy of knot detections, recall assesses the proportion of correctly detected knots, and mAP@0.5 provides an overall metric for detection accuracy at a threshold of IoU = 0.5.

\begin{figure*}[t!]
  \centering
  \begin{subfigure}[b]{0.45\textwidth} 
    \includegraphics[width=\textwidth]{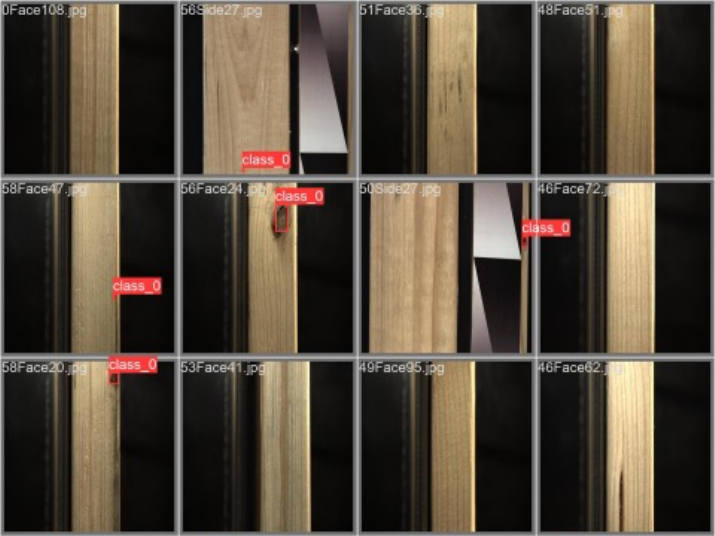}
    \caption{Annotated Ground Truth.}
    \label{fig:val_labels}
  \end{subfigure}
  \hspace{5pt} 
  \begin{subfigure}[b]{0.45\textwidth} 
    \includegraphics[width=\textwidth]{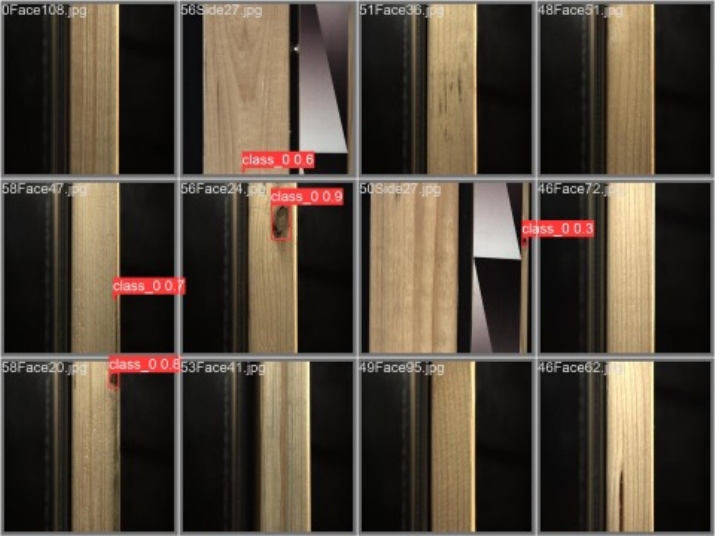}
    \caption{Knot Detection by YOLOv8l.}
    \label{fig:val_pred}
  \end{subfigure}
  \caption{Comparison of ground truth and model prediction results.}
  \label{fig:labels_pred}
\end{figure*}

\subsection{Knots Pairing}

\textbf{Knot Pair Dataset.} The knot pairing ground truth dataset consists of 62 wooden boards, 893 knots, and 451 knot pairs. All models were trained on triplets generated from the normalized dataset. To ensure robust evaluation, the dataset was split into training, validation, and test sets using an 8:1:1 ratio.

\textbf{Model Dataset.} We constructed the triplet network architectures and clustering strategies for the knot pairing as below:
\begin{itemize}
    \item \textbf{Standard Triplet Network.}  
    The Standard Triplet Network consists of an input layer followed by six fully connected layers. The first layer has 1024 units, and subsequent layers reduce the units to 512, 256, 128, 64, and a final 128-dimensional embedding layer. ReLU activation functions are applied after each layer, with dropout layers (rate = 0.3) added to the first four layers to prevent overfitting. The model was trained using the triplet loss function, which minimizes the distance between embeddings of similar knots while maximizing the distance between embeddings of dissimilar knots. The training was performed using the Adam optimizer with a learning rate of 0.0001 and a weight decay of $1 \times 10^{-5}$. 
    \item \textbf{SimCLR-based Network.}  
    The SimCLR-based network employs an encoder-projection structure. The encoder transforms input features into high-dimensional embeddings, while the projection head maps these embeddings into a lower-dimensional space. To enhance robustness, data augmentations including random noise addition, feature scaling, and feature dropping were applied during training. The model was trained using the NT-Xent (normalized temperature-scaled cross-entropy) loss function. The Adam optimizer with a learning rate of 0.001 was used, and the model was trained for 2,500 epochs with a batch size of 18. 

    \item \textbf{Triplet Network with Learnable Weights.}  
    In this variation, learnable weights were introduced for each feature in the dataset. During training, the network optimized these weights to dynamically adjust the importance of each feature, enhancing the accuracy of feature embeddings. The network architecture mirrored the standard triplet network, with the addition of a learnable weight layer applied before the first fully connected layer. The Adam optimizer was used with a learning rate of 0.0001 and a weight decay of $1 \times 10^{-5}$. The training spanned 2,000 epochs.

    \item \textbf{Triplet Network with Custom Weights.}  
    This variation applied fixed weights to the input features, leveraging domain knowledge to emphasize critical characteristics. For our dataset, which includes nine features (surface, \(d_{\text{min}}\), \(d_{\text{max}}\), \(k_1\) normalized, \(k_2\) normalized, longitudinal coordinate normalized, distance of knot center to bottom normalized, Knot type, and Pith Type), we assigned weights as [0.06, 0.06, 0.06, 0.06, 0.06, 0.06, 0.06, 0.18, 0.4]. These weights prioritize the "Knot type" and "Pith Type" features, as they are believed to significantly influence the accuracy of knot pairing. The architecture and training setup were identical to the Standard Triplet Network, with weights applied at the input layer. The Adam optimizer with the same hyperparameters was used, and training spanned 2,000 epochs. 
\end{itemize}

\begin{table}[t]
\centering
\caption{Performance metrics and training details for YOLO models. Arrows (\(\uparrow\), \(\downarrow\)) indicate whether greater or smaller values are better.}
\label{tab:yolo_results}
\begin{tabular}{|l|c|c|c|}
\hline
\textbf{Metric} & \textbf{YOLOv8n} & \textbf{YOLOv8m} & \textbf{YOLOv8l} \\ \hline \hline
\textbf{Precision (IoU\textgreater0.6)} (\(\uparrow\)) & 0.768            & 0.885            & 0.803            \\ \hline
\textbf{Recall} (\(\uparrow\))              & 0.831            & 0.746            & 0.873            \\ \hline
\textbf{mAP@0.5} (\(\uparrow\))            & 0.846            & 0.855            & 0.887            \\ \hline
\textbf{mAP@0.5:0.95} (\(\uparrow\))        & 0.521            & 0.485            & 0.500              \\ \hline
\hline
\textbf{Training Time (hrs)} (\(\downarrow\)) & 0.275            & 1.018            & 1.651            \\ \hline
\textbf{Model Size (GFLOPs)} (\(\downarrow\)) & 8.1              & 78.7             & 164.8            \\ \hline
\end{tabular}
\end{table}

To optimize clustering, we conducted a systematic search for the optimal distance threshold for each clustering model. Using trained networks, feature embeddings were first generated for all knots in the dataset, capturing their unique characteristics. Next, we performed a threshold search, where pairwise Euclidean distances were computed to form a distance matrix. A range of thresholds from 0.1 to 100 (step size = 0.01) was tested, and for each threshold, knots were grouped into clusters using Distance Threshold Clustering.
Finally, clustering accuracy was evaluated by comparing the predicted clusters to ground truth groupings, where knots with the same knot number under the same specimen were considered part of the same cluster. For example, if the ground truth pairs for a specimen were [1,2], [3,4], and [5,6], but the model produced [1,2], [3], and [4,5,6], the accuracy was calculated as 1/3. The optimal threshold was determined as the one yielding the highest clustering accuracy.

\section{Results}
\subsection{Transfer Learning with YOLO for Knot Detection}

We evaluated the performance of YOLOv8 models (YOLOv8n, YOLOv8m, YOLOv8l) on the knot detection task using precision, recall, F1 score, and mAP as metrics. 

YOLOv8l demonstrated the best overall detection capability, achieving the highest mAP@0.5 of 0.887 and consistently outperforming the other models across various evaluation metrics. YOLOv8m followed closely with a mAP@0.5 of 0.855, striking a balance between precision and processing efficiency. YOLOv8n, while being the fastest model with a training time of just 0.275 hours, achieved a slightly lower mAP@0.5 of 0.846. These results highlight the trade-offs between model complexity, accuracy, and computational efficiency.

The effectiveness of YOLOv8 models is visually highlighted in the comparison between the annotated ground truth (Figure~\ref{fig:val_labels}) and the model predictions (Figure~\ref{fig:val_pred}). 
These visual examples illustrate the accuracy of knot detection and demonstrate that the models adapt well to the characteristics of the wooden boards.

\begin{figure}[t]
  \centering
  \includegraphics[width=0.4\textwidth]{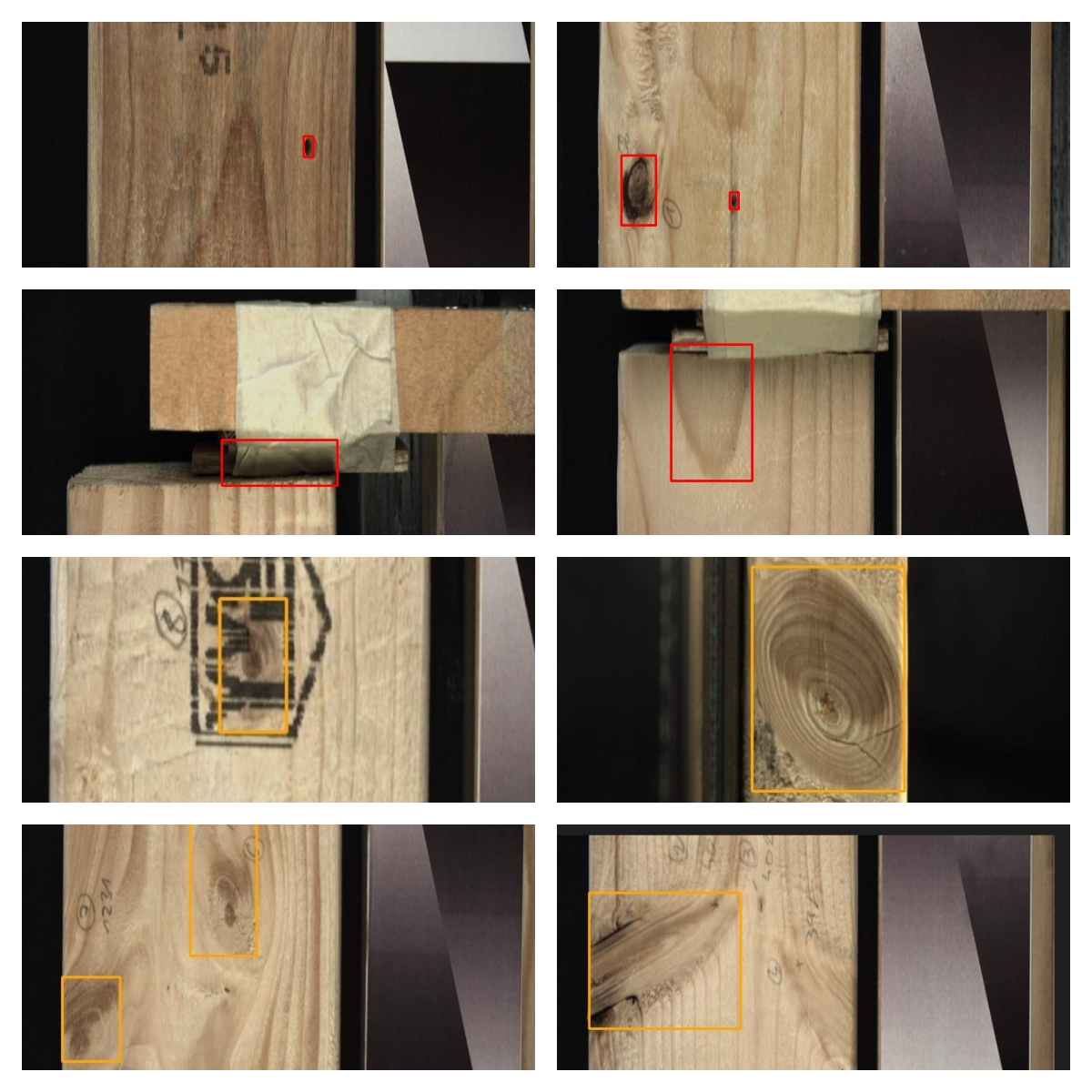}
  \caption{Error Analysis Preview: Red boxes indicate Detection Failures and orange boxes indicate Precision Issues.}
  \label{fig:error_analysis}
\end{figure}

\textbf{Error Analysis.}
To evaluate the performance of our model in the specific domain of wood processing, a thorough error analysis was conducted. This analysis helped categorize detection errors, providing insights for model refinement. In future industrial applications, where environmental factors such as factory lighting and camera exposure may impact performance, this module will play a crucial role in detecting and addressing model shortcomings.
Our error analysis focused on two primary types of detection errors:
\begin{enumerate}
\item \textbf{Detection Failures}: Cases where the model either fails to detect a knot (missed detection) or predicts bounding boxes that do not overlap with any ground truth annotation (false positives).
\item \textbf{Precision Issues}: Cases where the predicted bounding box overlaps with the ground truth but has an Intersection over Union (IoU) below 0.6, indicating insufficient localization accuracy.
 \end{enumerate}



We conducted an error analysis on the dataset used to train our model. These images were processed using the YOLOv8l model, and 708 annotated knots are included in the analysis. During this analysis, we identified 18 instances of Detection Failures and 88 instances of Precision Issues. A consolidated visual representation of these errors is shown in Figure~\ref{fig:error_analysis}, with red boxes highlighting detection failures and orange boxes indicating precision issues.
Upon closer examination, we observed the following patterns. The detection failures primarily occur when the model fails to recognize smaller knots and those with darker coloration. The model might have compromised on robustness, mistaking darker knots for shadows rather than actual knots. 
Additionally, shadows on the wooden boards during image capture and certain elliptical tree textures resembling knots were often incorrectly identified as knots.
Additionally, precision issues were prevalent when the model struggles with particularly large knots, knots with very light coloration, and those with unique or irregular boundary shapes. Additionally, knots partially obscured by printed text on the boards also result in lower IoU scores.
These observations suggest that our model had made certain trade-offs, favoring robustness by rejecting lower-confidence predictions. To further enhance the model’s performance, especially in these identified areas of weakness, targeted training with a specific focus on these challenging cases is necessary.

\subsection{Tagging Knots Pairs}

\begin{figure}[t]
  \centering
  \includegraphics[width=\linewidth]{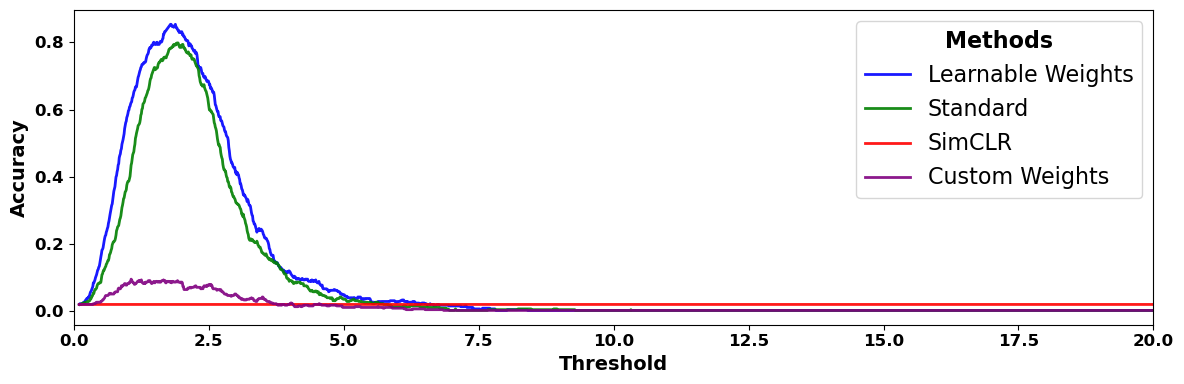}
  \caption{Models Performance with Threshold: Comparison of the standard Triplet Network, Triplet Network with Learnable Weights, Triplet Network with Custom Weights, and SimCLR-based method.}
  \label{fig:threshold_selection}
\end{figure}

We first obtained the optimal threshold for each model. Figure ~\ref{fig:threshold_selection} shows the performance of four models for different threshold selections.
The performance metrics for each model are summarized in Table~\ref{tab:models_performance_thresholds}.
The standard triplet network achieves an accuracy of 0.8 at a threshold of 1.9. In contrast, the triplet network with learnable weights performs better, reaching an accuracy of 0.85 at a threshold of 1.79. The triplet network with custom weights shows significantly lower performance, achieving an accuracy of only 0.1 at a threshold of 1.05. Similarly, the SimCLR-based method performs poorly, with an accuracy of 0.02 at a threshold of 0.1.

\begin{table}[t]
\centering
\caption{Performance Metrics of Different Models; This table highlights the varying effectiveness of the different models, with the Triplet Network with Learnable Weights showing the highest accuracy.}
\label{tab:models_performance_thresholds}
\begin{tabular}{|c|c|c|}
\hline
\textbf{Model} & \textbf{\makecell[c]{Optimal \\ Threshold}} & \textbf{Accuracy (\(\uparrow\))} \\ \hline
\hline
Standard Triplet Network & 1.9 & 0.8 \\ \hline
Triplet Network with Learnable Weights & 1.79 & 0.85 \\ \hline
Triplet Network with Custom Weights & 1.05 & 0.1 \\ \hline
SimCLR-based Method & 0.1 & 0.02 \\ \hline
\end{tabular}
\end{table}

\begin{figure}[t]
  \centering
  \begin{minipage}[b]{0.48\textwidth}
    \includegraphics[width=\textwidth]{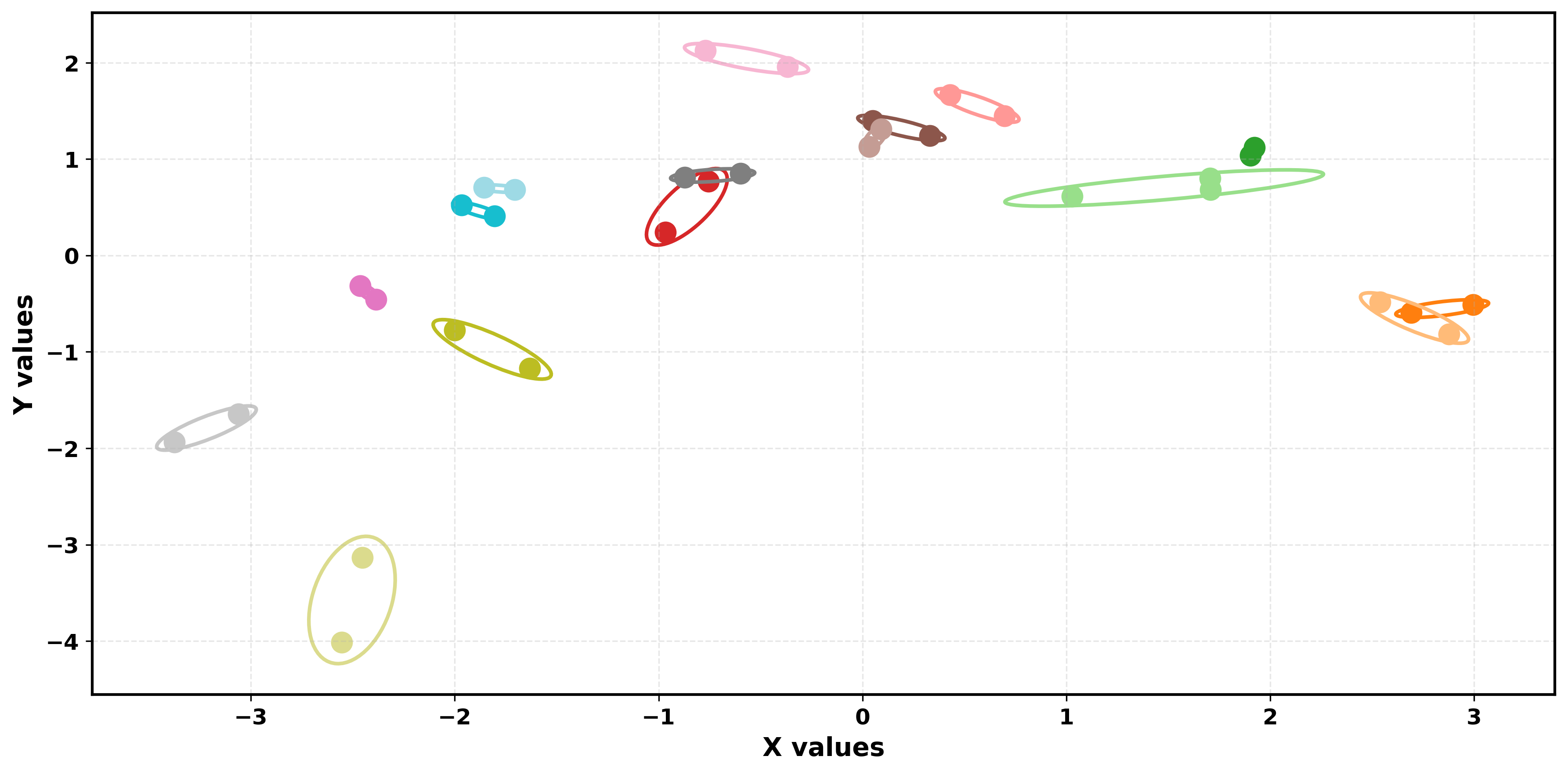}
    \caption{A two-dimension visualization of clustering.}
    \label{fig:clusters}
  \end{minipage}
\end{figure}

\textbf{Visualization.}  
To further visually demonstrate the practicability of our method, the high-dimensional embeddings generated by the model are visualized using PCA, mapping 128-dimensional embeddings to 2 dimensions. As shown in Figure~\ref{fig:clusters}, dots of the same color represent knots in the same cluster (i.e. pairs). The close proximity of dots within clusters highlights the effectiveness and meaningfulness of the proposed approach.

\begin{table}[t]
\centering
\caption{Feature Weights from Triplet Network with Learnable Weights}
\label{tab:feature_weights}
\begin{tabular}{|c|c|}
\hline
\textbf{Feature} & \textbf{Weight} \\ \hline
\hline
Surface & 0.098 \\ \hline
d\_min & 0.040 \\ \hline
d\_max & 0.037 \\ \hline
k1 & 0.153 \\ \hline
k2 & 0.159 \\ \hline
Longitudinal Coordinate & 0.231 \\ \hline
Distance of knot center to bottom & 0.122 \\ \hline
Knot Type & 0.111 \\ \hline
Pith Type & 0.050 \\ \hline
\end{tabular}
\end{table}

\textbf{Analysis and Discussion.}
We extracted the weight matrix from the trained triplet network using the learnable weights model, which offers insights into the significance of each feature in predicting knot pairing accuracy. 
A summary of the weights and feature importance is provided in Table~\ref{tab:feature_weights}. It also shows that the features k1, k2, and longitudinal coordinates play a crucial role in achieving high knot pairing accuracy. Contrary to our initial intuition, the knot and pith types, which we assumed to be more significant, have less weight.

This finding is further reflected in the performance of the triplet network with custom weights. When we manually assign higher weights to knot type and pith type (18\% and 40\% respectively), the model achieves a significantly lower accuracy of only 0.1 after generalization. This demonstrates that our manual weighting approach is not aligned with the actual feature importance learned by the model.
Similarly, the SimCLR-based model exhibits a low accuracy, indicating that perturbations and disturbances in any feature can substantially impact the effectiveness of knot pairing. This suggests that all features contribute to the knot pairing task, and maintaining high feature accuracy and purity is critical for optimal performance.

In summary, our analysis reveals that the features k1, k2, and longitudinal coordinates are the most critical for the knot pairing task. These features significantly influence the model's ability to pair knots accurately. Moreover, high-quality data is crucial in achieving better performance in knot pairing. 



\section{Conclusion}
\label{sec:conclusion}
This study presents a scalable and automated approach for knot detection and pairing in timber processing, leveraging machine learning techniques to address the limitations of traditional manual inspection. By integrating computer vision-based detection with feature-driven pairing methodologies, we demonstrate the feasibility of AI in wood analysis and its potential for industrial applications.
In the detection stage, high-resolution surface images of wooden boards were collected and annotated to train YOLOv8 models via transfer learning. Our results show that YOLOv8l achieved an mAP@0.5 of 0.887, effectively capturing intricate knot features, while the lightweight YOLOv8n offered a computationally efficient alternative with minimal accuracy trade-offs.
For knot pairing, detected knots were analyzed based on multidimensional feature extraction, with a triplet neural network mapping features into a latent space for clustering. The Triplet Network with Learnable Weights achieved a pairing accuracy of 0.85, where k1, k2, and longitudinal coordinates playing crucial roles in accurate pairing. 
These findings provide valuable insights into feature significance in wood analysis and contribute to advancing automation in the timber industry.

\bibliographystyle{IEEEtran}
\bibliography{references}

\end{document}